\documentclass[runningheads]{llncs}
\usepackage[T1]{fontenc}
\usepackage[utf8]{inputenc}

\usepackage[T1]{fontenc}
\usepackage{amsmath}
\usepackage{amsfonts}
\usepackage{multirow}
\usepackage{makecell}
\usepackage{pifont}
\usepackage{booktabs}

\newcommand{\xmark}{\ding{55}}
\usepackage{graphicx,verbatim}
\begin{document}
\title{OpenRC: An Open-Source Robotic Colonoscopy Framework for Multimodal Data Acquisition and Autonomy Research}
\titlerunning{OpenRC: Open-Source Robotic Colonoscopy Framework}
%
\author{Siddhartha Kapuria\inst{1} \and
Mohammad Rafiee Javazm\inst{1} \and
Naruhiko Ikoma\inst{2} \and 
Joga Ivatury\inst{3} \and Mohammad Ali Nasseri\inst{4} \and Nassir Navab\inst{4} \and Farshid Alambeigi\inst{1} }

\authorrunning{S. Kapuria et al.}

\institute{Walker Department of Mechanical Engineering, The University of Texas at Austin, Austin, 78712, TX, USA \and
Department of Surgical Oncology, Division of Surgery, The University of Texas MD Anderson Cancer Center, Houston, 77030, TX, USA \and
Department of Surgery \& Perioperative Care, Dell Medical School, The University of
Texas at Austin, Austin, 78712, TX, USA \and
School of Medicine and Health, Technical University of Munich, Munich, 80333, Germany
}

\maketitle              
\begin{abstract}

Colorectal cancer screening critically depends on colonoscopy, yet existing platforms offer limited support for systematically studying the coupled dynamics of operator control, instrument motion, and visual feedback. This gap restricts reproducible closed-loop research in robotic colonoscopy, medical imaging, and emerging vision-language-action (VLA) learning paradigms. To address this challenge, we present OpenRC, an open-source modular robotic colonoscopy framework that retrofits conventional scopes while preserving clinical workflow. The framework supports simultaneous recording of video, operator commands, actuation state, and distal tip pose. We experimentally validated motion consistency and quantified cross-modal latency across sensing streams. Using this platform, we collected a multimodal dataset comprising 1,894 teleoperated episodes ($\sim$19 hours) across 10 structured task variations of routine navigation, failure events, and recovery behaviors. By unifying open hardware and an aligned multimodal dataset, OpenRC provides a reproducible foundation for research in multimodal robotic colonoscopy and surgical autonomy.


\end{abstract}

\section{Introduction}
\label{sec:intro}

Colorectal cancer (CRC) is a major global health burden, ranking as the third most common cancer worldwide, with 1.9 million new cases and 935,000 deaths in 2020 \cite{Sung2021GlobalCS,xi2021global}. Early detection is critical, as survival strongly depends on stage of cancer. Although colonoscopy is the gold standard and reduces mortality through removal of precancerous lesions \cite{Jemal2002CancerS2}, it still has adenoma miss rates of up to 34\% \cite{Zhao2019MagnitudeRF}. These limitations arise from optical challenges (e.g., occlusion, blur), operator variability, and device ergonomics  \cite{Ko2010ComplicationsOC}. 

Investigating the literature, prior efforts in addressing colonoscopy challenges can broadly be categorized into two largely independent directions. The first emphasizes robotic and hardware innovation, including robotic  platforms and motorized retrofits designed to improve dexterity, stability, and ergonomics  (e.g., 
\cite{Kume2015EndoscopicRobot,Basha2023GenericScope,Lee2020EasyEndo,bovskoski2021robotics}). The second group focuses on AI-driven assistance through perception, classification, and SLAM, solely relying on colonoscopic video feed without considering instrument kinematics and control (e.g.,  \cite{nie2024review,tham2023imitation,ozyoruk2021endoslam}). Consequently, perception and control in these groups of efforts are typically studied in isolation. 

Emerging autonomy-driven frameworks, particularly Vision-Language-Action (VLA) systems, require joint access to observations, actions, and system state. While such multimodal datasets are standard in broader robotics domains \cite{open_x_embodiment_rt_x_2023,walke2023bridgedata}, comparable resources remain limited in colonoscopy. 
Existing open source datasets predominantly provide video data and, in some cases, distal pose measurements, but rarely capture synchronized operator actions or internal actuation state (Table \ref{tab:dataset_comparison}). Furthermore, there is currently no unified open-source hardware and software platform enabling reproducible closed-loop experimentation in colonoscopy. This lack of integrated platforms and datasets presents a fundamental barrier to reproducible research in robotic assistance and autonomy. 








\begin{table}[t]
\caption{Comparison of colonoscopy datasets and resources by available modalities}
\label{tab:dataset_comparison}
\centering
\scriptsize
\setlength{\tabcolsep}{3pt}
\renewcommand{\arraystretch}{1.2}
\begin{tabular}{
p{3.5cm}
>{\centering\arraybackslash}p{0.6cm}
>{\centering\arraybackslash}p{1.2cm}
>{\centering\arraybackslash}p{1.2cm}
>{\centering\arraybackslash}p{1.2cm}
>{\centering\arraybackslash}p{1.6cm}
>{\centering\arraybackslash}p{1.5cm}
}
\toprule
\textbf{Dataset }& \textbf{Video} & \textbf{\shortstack[c]{Operator\\Actions}} &
\textbf{\shortstack[c]{Actuation\\State}} &
\textbf{\shortstack[c]{6DoF\\Pose}} &
\textbf{\shortstack[c]{Cross-Modal\\Alignment}} &
\textbf{\shortstack[c]{Open\\Hardware}} \\
\midrule

Clinical image/video datasets \cite{bernal2015wm,jha2020kvasir,Borgli2020,Biffi2024REALColonAD}
& \checkmark & \xmark & \xmark & \xmark & \xmark & \xmark \\
\hline
SimCol / SimCol3D \cite{rau2023bimodal,rau2024simcol3d}
& \checkmark & \xmark & \xmark & \checkmark\,(synthetic) & \checkmark & \xmark \\
\hline
EndoSLAM \cite{ozyoruk2021endoslam}
& \checkmark & \xmark & \xmark & \checkmark & \checkmark & \xmark \\
\hline
EndoMapper \cite{Azagra2022EndomapperDO}
& \checkmark & \xmark & \xmark & \checkmark\,(simulated subset) & \checkmark & \xmark \\
\hline
C3VDv2 \cite{golhar2025c3vdv2}
& \checkmark & \xmark & \xmark & \checkmark & \checkmark & Phantom STL/CAD \\
\hline
Copenhagen Colonoscopy Coordinate DB \cite{Cold2025MappingTC}
& \xmark & \xmark & \xmark & \checkmark & Single-modality & \xmark \\
\hline
\textbf{OpenRC (Ours)}
& \checkmark & \checkmark & \checkmark & \checkmark & \checkmark & \checkmark \\

\bottomrule
\end{tabular}
\end{table}

To address these challenges, we introduce \textit{OpenRC}, \textit{an open-source robotic colonoscopy research framework} for reproducible closed-loop experimentation and unified multimodal data collection. This low-cost platform retrofits conventional colonoscopes with modular actuation while preserving clinical workflow and synchronizes operator commands, actuation state, electromagnetic (EM) -tracked distal tip pose, and video. 
Using this system, we conducted comprehensive hardware and data acquisition validation studies, and collected a multimodal dataset comprising 1,894 teleoperated episodes ($\sim$19 hours)  spanning navigation and failure–recovery scenarios. \textit{Both the hardware design and dataset will be made publicly available upon acceptance to support reproducible research in surgical autonomy, control-aware perception, and multimodal learning.}

\begin{figure}[t]
    \centering
    \includegraphics[width=0.95\textwidth]{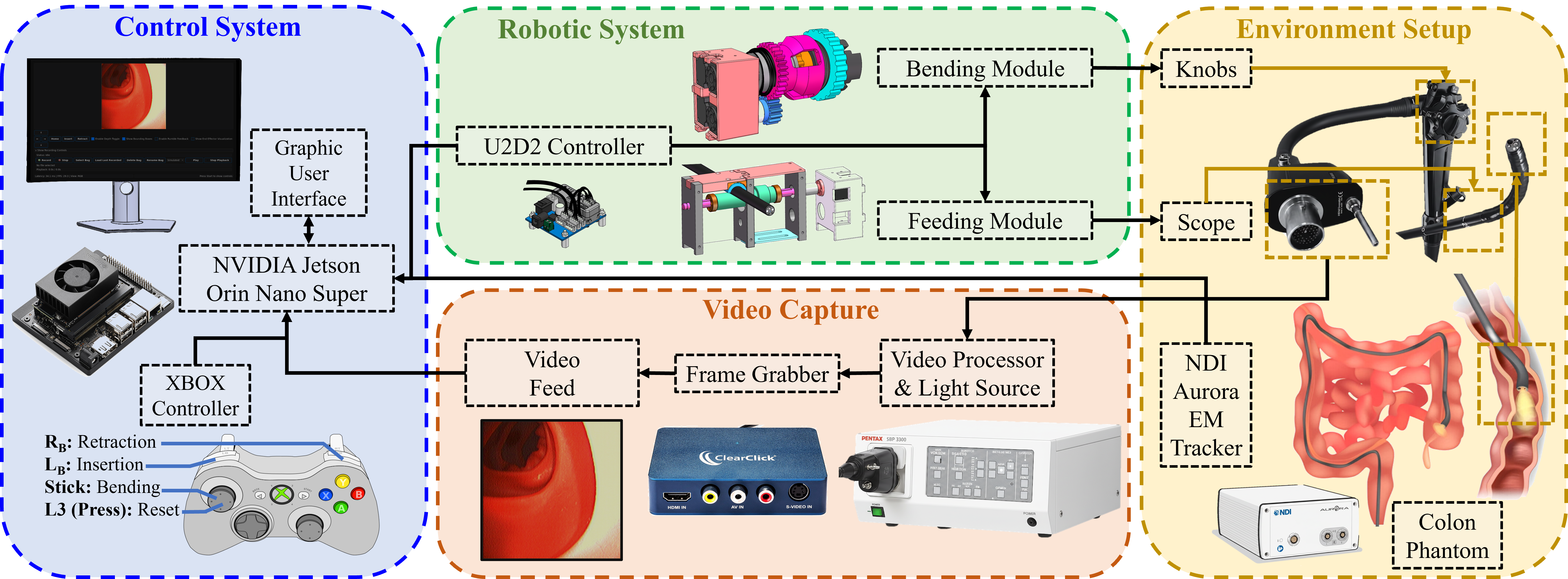}
    \caption{Overview of the proposed OpenRC framework, including hardware interfaces, actuation, data acquisition, and sensing via EM tracking.}
    \label{fig:system}
    
\end{figure}

\section{Methods}

As shown in Fig.~\ref{fig:system}, to support research in computer vision, surgical autonomy, control-aware perception, and multimodal learning for colonoscopy procedures, the proposed open-source platform comprises four main modules: (1) a retrofitted robotic actuation unit for conventional colonoscopes; (2) integrated sensing and synchronized data logging; (3) configurable phantom testbeds; and (4) a ROS-based software architecture with a unified user interface.  The entire framework (excluding the EM tracker) can be assembled for under \$5,000 USD, promoting accessibility and reproducibility. 
 \textit{Of note, a full bill of materials and fabrication procedure for the robot and phantom test bed will be made available post-acceptance.} The following sections will describe these modules in detail.

\subsection{Robotic Platform for Colonoscopy}
The OpenRC system is designed to retrofit conventional commercially-available colonoscopes, allowing for the development of a low-cost robotic system with repeated actuation and synchronized sensing without modifying the original device. In particular, it targets the three clinically relevant degrees-of-freedom (DoFs): insertion/retraction and distal-tip bending in the lateral and vertical directions.  
Without loss of generality, for this study, we use a PENTAX EC-3840LK colonoscope (Pentax Medical, Japan) as a representative model, although the system is designed to be compatible with other colonoscopes. 

\begin{figure}[t]
    \centering
    \includegraphics[width=0.98\linewidth]{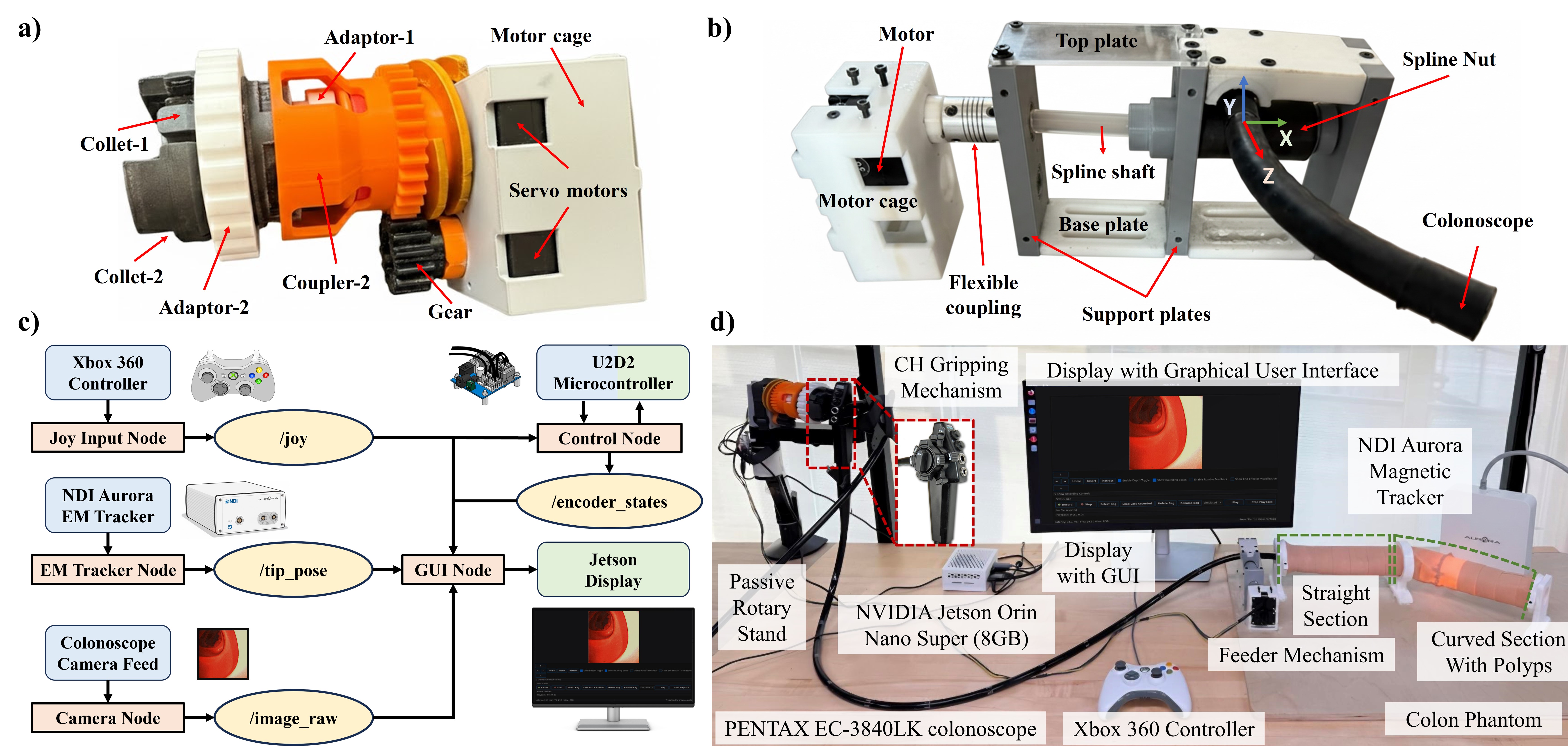}
    \caption{OpenRC components: (a) bending module, (b) feeding module, (c) ROS~2 graph showing data streams, and (d) experimental setup for data collection.
    }
    \label{fig:components}
\end{figure}

\textbf{Bending Module.}
\label{sec:bending_module}
As shown in Fig.~\ref{fig:components}(a), the bending DoFs are motorized using two nested 3D-printed collets that transmit torque directly to the existing control-handle (CH) knobs without modifying the colonoscope. 
Inspired by a drill collet, the concentric multi-jaw design enables secure gripping and rapid attachment/detachment. Two DYNAMIXEL \texttt{XM540-W270-R} servos (1{:}270 gearbox, 12-bit absolute encoder; 0.088$^\circ$ output resolution) provide actuation, delivering integrated sensing and sufficient holding torque, eliminating the need for the native lock. An additional 1{:}2 reduction in the outer collet improves command resolution. Components are fabricated from nylon–carbon fiber composite and ABS plastic to balance strength and manufacturability.

\textbf{Feeding Module.}
As shown in Fig.~\ref{fig:components}(b), the feeding module uses a compact friction-based drive for insertion and retraction of the scope. A motor-driven urethane-sleeved feeder roller is used to provide traction, while a low-friction idler-ball guide applies an adjustable normal force via preload screws, ensuring consistent and controllable grip. The applied pressure can be tuned to accommodate different colonoscope diameters and to balance reliable advancement with safe force application. The roller is actuated using a DYNAMIXEL \texttt{XM430-W350-R} servo (1{:}350 gearbox, 12-bit absolute encoder; 0.088$^\circ$ resolution).  The feeding unit is mechanically independent of the bending module.

\subsection{Software Architecture and Control Interface}
\label{sec:software}
As shown in Fig. \ref{fig:components}(c), the system is built on a compact ROS~2 (Humble) stack running on an embedded NVIDIA Jetson Orin Nano Super (8\,GB). Commands for bending and insertion actuation are issued to the DYNAMIXEL servos over an RS-485 daisy chain (ROBOTIS, South Korea) via a U2D2 interface and power hub board. Operator inputs are provided through an Xbox 360 controller and mapped to an action vector $(\texttt{bend\_x}, \texttt{bend\_y}, \texttt{insertion}, \texttt{home})$ normalized to $[-1, 1]$, enabling simultaneous 3-DoF control and a reset-to-home command. The modular ROS~2 design enables straightforward integration of additional sensing modalities or autonomous control policies. 
Teleoperation and \texttt{rosbag} logging execute in parallel to preserve control timing and ensure lossless recording.

\subsection{Sensing and Logged Modalities}
\label{sec:sensing}

Through the ROS-based architecture described in Sec. \ref{sec:software}, the OpenRC platform records multiple sensing modalities during colonoscope navigation via \texttt{rosbag},  timestamped using a shared system clock. The following streams can be recorded:

\textbf{Colonoscope Video.}
Video is captured at 30 frames per second at 383 $\times$ 396 pixels (native resolution) using the inbuilt camera at the distal tip of the colonoscope. As shown in Fig. \ref{fig:system}, this feed is streamed via a Video to USB 1080P Capture device (ClearClick, USA) connected to a Video Processor and Light Source EPM 3300 (Pentax Medical, Japan) attached to the colonoscope.

\textbf{Operator Commands.}
Operator intent is recorded as a four-dimensional action vector as described in Sec. \ref{sec:software}
, at a rate of 50 Hz. The normalized inputs provide a compact, abstracted representation of high-level operator commands independent of controller button mappings. 

\textbf{Robot State.}
In addition to operator commands, the joint-space actuation states are recorded at 50 Hz from the motor controllers. These  signals comprise encoder-derived measurements for dual-axis distal bending and longitudinal translation, normalized to output shaft degrees-of-rotation to decouple encoder resolution from physical actuation. Importantly, due to the device’s inherent flexibility and environmental coupling, this joint-space representation does not uniquely map to the task-space distal tip pose.


\textbf{Distal Tip Pose.}
Distal tip pose is measured at 6-DoF using an EM tracking system (NDI Aurora, Canada), represented as Cartesian position ($m$) and quaternion orientation. Since the colonoscope's inherent flexibility and environmental interaction preclude a fixed spatial calibration, pose data are reported directly in the tracker's coordinate frame. 
\subsection{Phantom Testbeds}
\label{sec:testbed}

We used two flexible colon phantoms: (i) a custom-built silicone phantom with embedded polyps and (ii) a commercially available training phantom.  
Importantly, the phantom-based setting enables acquisition of precise 6-DoF ground truth, which is not always feasible in clinical practice.  Using both phantoms increases variation in geometry, surface appearance, and deformation characteristics without changing the robotic actuation and logging stack, improving the diversity of the resulting multimodal dataset.

\textbf{Custom polyp phantom.}
Our custom phantom was designed to support both navigation and lesion-observation studies by permanently embedding realistic silicone polyps along the lumen. In brief, polyps were designed following the Paris classification \cite{Kaltenbach2020EndoscopicRO,Kara2023ARA}, cast in silicone using 3D-printed molds, and integrated into a modular silicone colon segment during fabrication. The phantom diameter was selected based on reported average human colon dimensions \cite{Alazmani2016QuantitativeAO}. Multiple segments can be joined to adjust overall length and curvature. 

\textbf{Commercial flexible phantom.}
We also collected data in a commercially available flexible colonoscopy training phantom (Kyoto Kagaku, Japan).

\begin{figure}[t]
    \centering
    \includegraphics[width=0.98\textwidth]{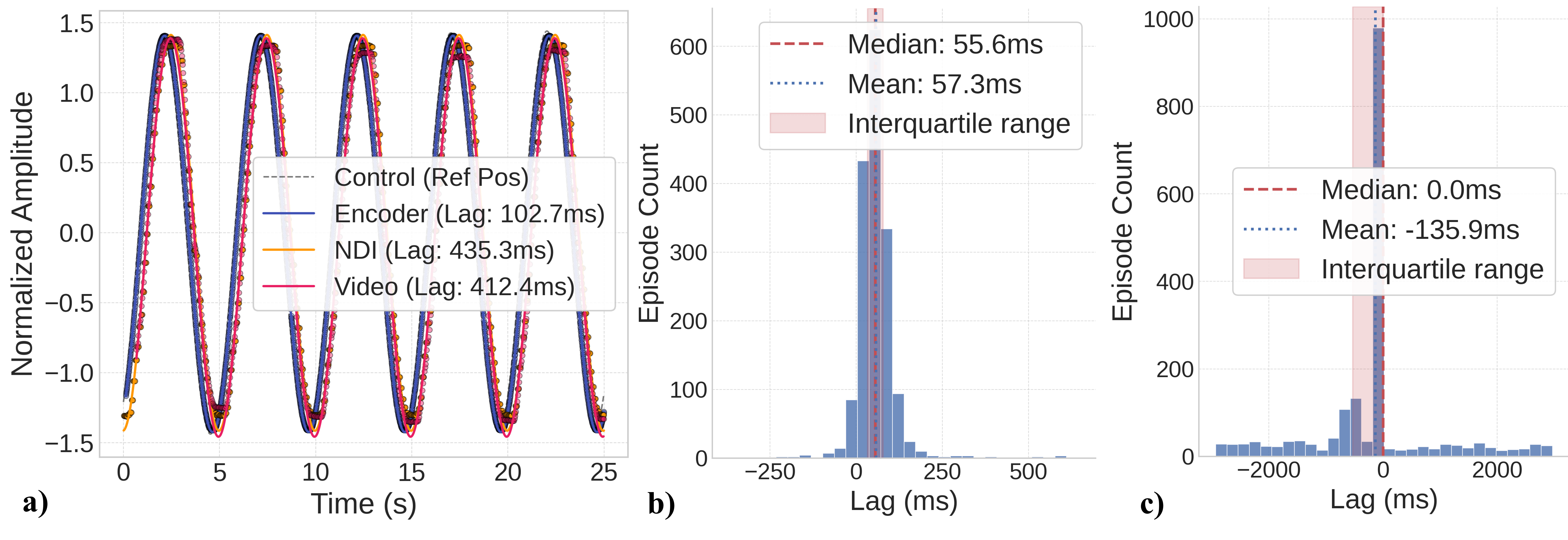}
    \caption{Results for characterization and synchronization showing: (a) sinusoidal system response characterization, and histograms of estimated residual lag distributions for (b) Operator Action vs State, and (c) State vs Tip Position.}
    \label{fig:calibration}
    
\end{figure}

\section{System Characterization}
\label{sec:calibration}

As described in Sec.~\ref{sec:sensing}, OpenRC integrates operator input, joint-space motor states, EM-based distal tip tracking, and colonoscope video, which operate through independent acquisition pipelines. These readings are not inherently time-synchronized. Therefore, to assess hardware motion consistency and quantify cross-modal latency, we  performed  controlled excitation experiments. Specifically, we applied a low-frequency sinusoidal command (0.2\,Hz) along the X axis (see Fig. \ref{fig:components}(b)) for bending. This excitation produced smooth, periodic motion observable across control commands, motor encoder measurements, distal tip position reported by the EM tracker, and motion estimates derived from optical flow (computed via the Lucas-Kanade method \cite{lucas1981iterative} and averaged across 5 manually selected representative keypoints in the video frame). For each modality, a sinusoidal model was fitted to the measured response, and relative phase differences were computed via least-squares regression. As shown in Fig.~\ref{fig:calibration}(a), the measured responses closely follow the commanded input, indicating consistent actuation. The estimated temporal offsets relative to the control action are approximately 102 ms for motor encoder state, 435 ms for EM tracking, and 412 ms for optical-flow-derived motion. These differences in offsets are consistent with the external data acquisition pipelines for each modality. 
\section{Multimodal Robotic Colonoscopy Dataset}
\label{sec:dataset}

Using the OpenRC framework, we collected a comprehensive multimodal dataset that captures data streams across the modalities described in Sec. \ref{sec:sensing}. This section outlines the dataset structure, key features, technical validation, and potential applications in advancing robotic colonoscopy and autonomy research. To ensure accessibility, the dataset is saved in the LeRobot 2.1 format \cite{cadene2024lerobot}.

\begin{figure}[t]
    \centering
    \includegraphics[width=0.98\textwidth]{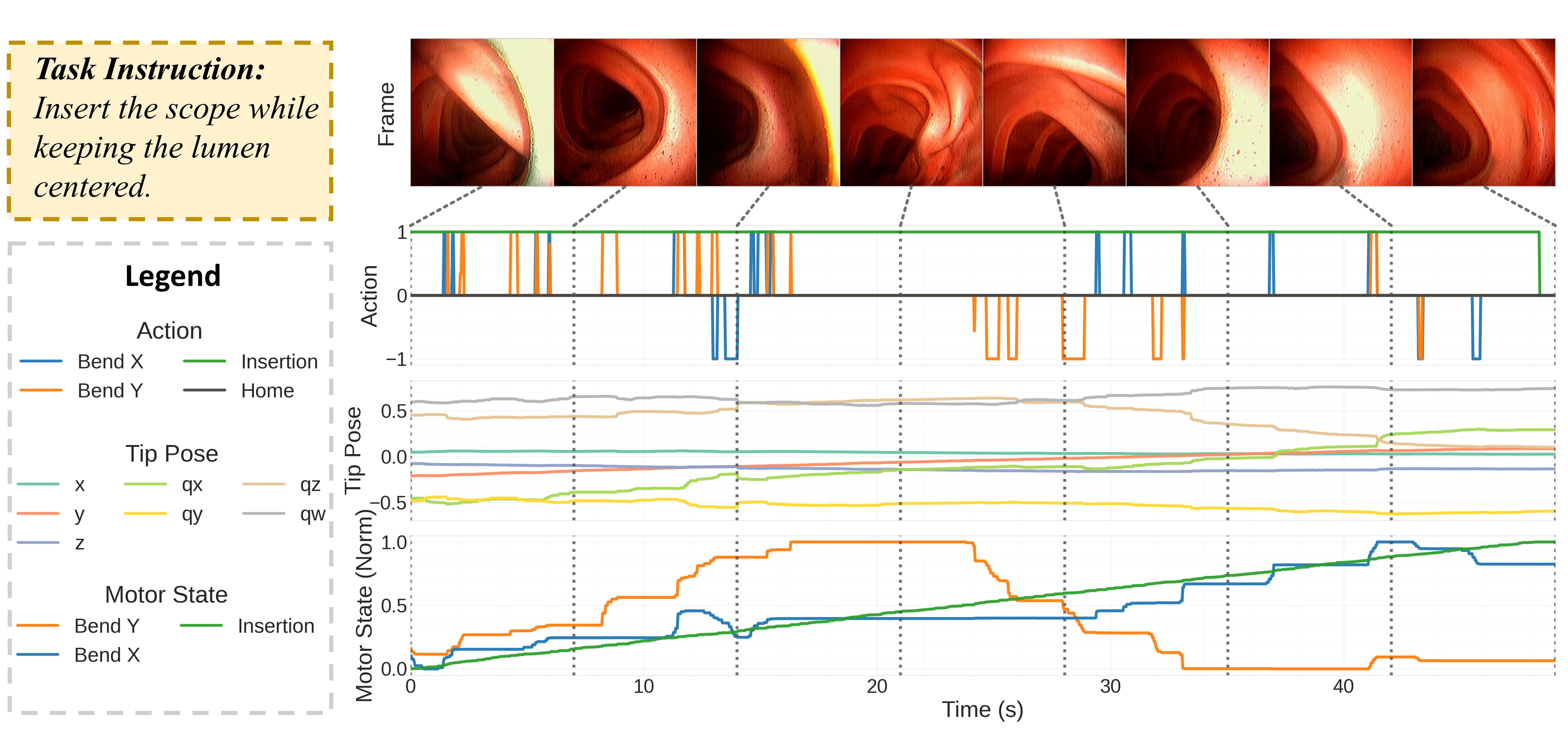}
    \caption{Example episode with synchronized multimodal recordings from the robotic colonoscopy dataset. From top to bottom:  colonoscope video snapshots sampled from the same time axis as data streams; operator control actions; distal tip pose; robot state (normalized to $[-1, 1]$ per axis for ease of visualization).
    }
    \label{fig:montage}
    
\end{figure}

\textbf{Episode Definition.}
The dataset is organized into episodes, where each episode corresponds to a single continuous teleoperated navigation trial performed within a colon phantom. 
Episodes include active manipulation, pauses, and adjustments, with durations varying by operator behavior and phantom geometry. For each episode, a short natural language instruction describing the intended task or navigation objective is recorded as metadata. These instructions are provided prior to teleoperation and remain fixed for the duration of the episode.  Fig. \ref{fig:montage} illustrates a representative teleoperated episode from the dataset for the insertion task, showing the temporal evolution across modalities. 

\textbf{Dataset Characteristics.}

\begin{figure}[t]
    \centering
    \includegraphics[width=0.98\textwidth]{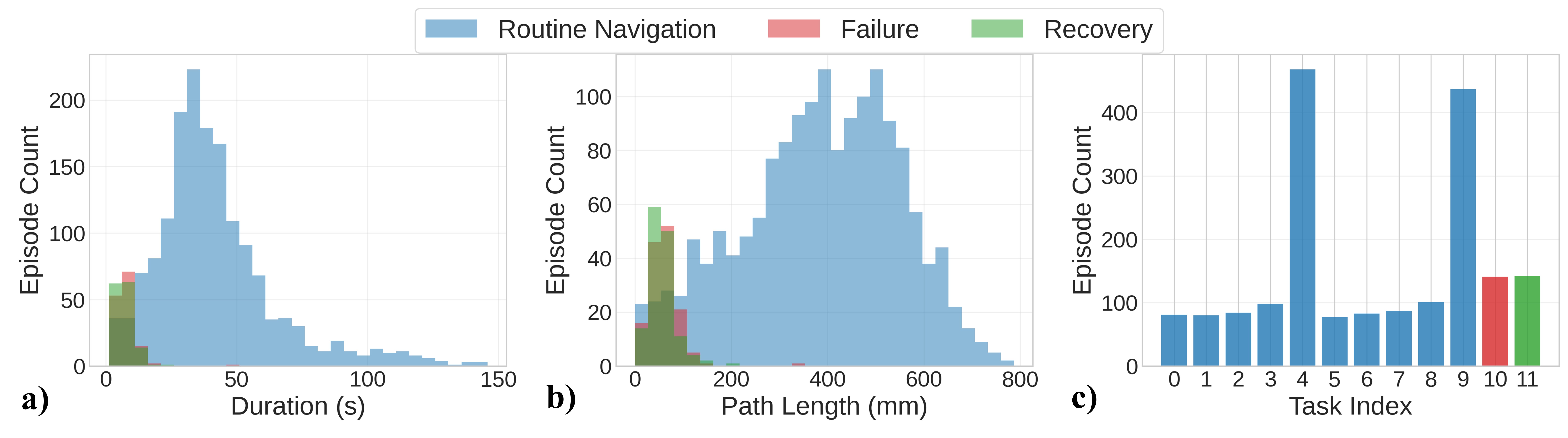}
    \caption{Episode-level characteristics of the OpenRC Dataset showing distributions of (a) episode duration, (b) trajectory length, and (c) recorded task. Detailed task descriptions will be provided in the dataset repository.}
    \label{fig:stats}
    
\end{figure}

The dataset comprises 1,894 episodes spanning $\sim$19 hours of teleoperated navigation across 10 task variations focused on insertion–retraction and structured wall-scanning behaviors. Specifically, tasks 0–4 correspond to \texttt{insertion} with structured scanning objectives: \texttt{bottom wall} (0), \texttt{left wall} (1), \texttt{right wall} (2), \texttt{top wall} (3), and \texttt{lumen-centered navigation} (4). Tasks 5–9 mirror these objectives during \texttt{retraction}. As shown in Fig.~\ref{fig:stats}(c), lumen-centered insertion and retraction tasks (tasks 4 and 9) predominate, reflecting their central role in scope manipulation, while wall-scanning tasks (tasks 0-3 and 5-7) are less frequent due to increased steering complexity. The dataset also includes 142 \texttt{failure} and 141 \texttt{recovery} episodes capturing common challenges such as lumen loss, wall contact, and fold engagement, along with corrective maneuvers. This enables structured study of error recovery in colonoscopic navigation.
Episode-level statistics in Fig.~\ref{fig:stats}(a–b) show a clear separation between routine navigation and failure/recovery segments. Navigation episodes are substantially longer and involve greater trajectory lengths, whereas failure and recovery segments are shorter and more localized in motion.

\textbf{Data Synchronization.}
To ensure accurate cross-modal analysis and support typical VLA learning pipelines, all data streams are resampled to $30$~Hz and calibrated using the offsets from Sec.~\ref{sec:calibration}, with the video feed as reference. To validate post-alignment stability, we estimated the residual temporal lag $\tau^*$ between modality pairs $(x, y)$, namely, operator action vs.\ actuation state, and actuation state vs.\ NDI tip pose. To reduce sensitivity to absolute offsets, we extracted normalized, velocity-like representations. For a given signal $x$ (and identically for $y$), we computed the finite difference Euclidean norm $v^x_t = \|x_t - x_{t-1}\|_2$, and applied min-max normalization as $\hat{v}^x = (v^x - \min(v^x)) / (\max(v^x) - \min(v^x))$. The residual offset was defined as the lag that maximizes the Pearson correlation coefficient $\rho$ over a bounded symmetric window $[-\tau_{\max}, \tau_{\max}]$, formulated as $\tau^* = \arg\max_{\tau} \rho ( \hat{v}^x_{t}, \hat{v}^y_{t+\tau} )$, where $\rho$ was computed over the valid temporal overlap of the two signals. As shown in Fig.~\ref{fig:calibration}(b-c), residual lag distributions show temporal stability across modalities. Action vs.\ state (Fig.~\ref{fig:calibration}(b)) shows a tightly clustered median offset of $55.6$~ms ($\sim$$1.6$ frames), well within acceptable limits for downstream policy learning. State vs.\ distal tip positions (Fig.~\ref{fig:calibration}(c)) center at $0.0$~ms, validating the baseline controller-to-tracker alignment. The wider variance and fat tails in this case can be attributed to the continuum dynamics of the deformable environment, which can decouple distal motion from actuation. 

\section{Discussion and Conclusion}
In this work, we introduced OpenRC, a low-cost ($<\$5,000$ USD) open-source robotic colonoscopy framework designed to retrofit conventional colonoscopes for reproducible closed-loop experimentation and unified multimodal data collection. The platform provides modular actuation of the clinically relevant DoFs, integrated multimodal sensing, and a ROS~2-based teleoperation and logging framework. Furthermore, we release a temporally synchronized multimodal data-set (in LeRobot 2.1 format) collected using this platform, comprising 1,894 episodes ($\sim$19 hours) of teleoperated procedures, including insertion, retraction, structured wall-scanning, and failure/recovery scenarios. Experimental validation confirmed reliable system actuation and yielded a stable post-alignment median residual lag of 55.6 ms between operator actions and actuation states, and 0.0 ms between motor states and EM-tracked distal tip poses. The collected synchronized action-state-pose-video trajectories provide a reproducible foundation for advancing research in surgical autonomy, control-aware perception, and multimodal learning in colonoscopy procedures. 

We acknowledge that the scope of our platform and dataset is currently limited by the controlled phantom environment used for data collection. Specifically, these ex vivo setups do not comprehensively capture patient-specific anatomical variability and dynamic physiological motion.   In the future, we plan to test the system on more realistic colon phantoms and animal models. We will also leverage this platform and dataset to train and evaluate robust VLA models for autonomous colonoscopic navigation and error recovery.

\bibliographystyle{splncs04}
\bibliography{mybib1}

\end{document}